\documentclass[runningheads]{llncs}

\usepackage[mobile]{eccv}

\usepackage{eccvabbrv}

\usepackage{graphicx}
\usepackage{booktabs}
\usepackage{array}
\usepackage[accsupp]{axessibility}  
\usepackage{tabularx}

\usepackage{hyperref}

\usepackage{orcidlink}

\begin{document}

\title{Generative Phomosaic with Structure-Aligned and Personalized Diffusion} 


\author{Jaeyoung Chung\thanks{indicates equal contribution.}\inst{1} \and Hyunjin Son\inst{\star1} \and Kyoung Mu Lee\inst{1}}

\authorrunning{J. Chung et al.}

\institute{Seoul National University, Seoul, South Korea \\
\email{\{robot0321, hyunjin.son, kyoungmu\}@snu.ac.kr} \\
}

\maketitle

\begin{center}
    \vspace{-4mm}
    \includegraphics[width=1.0\textwidth]{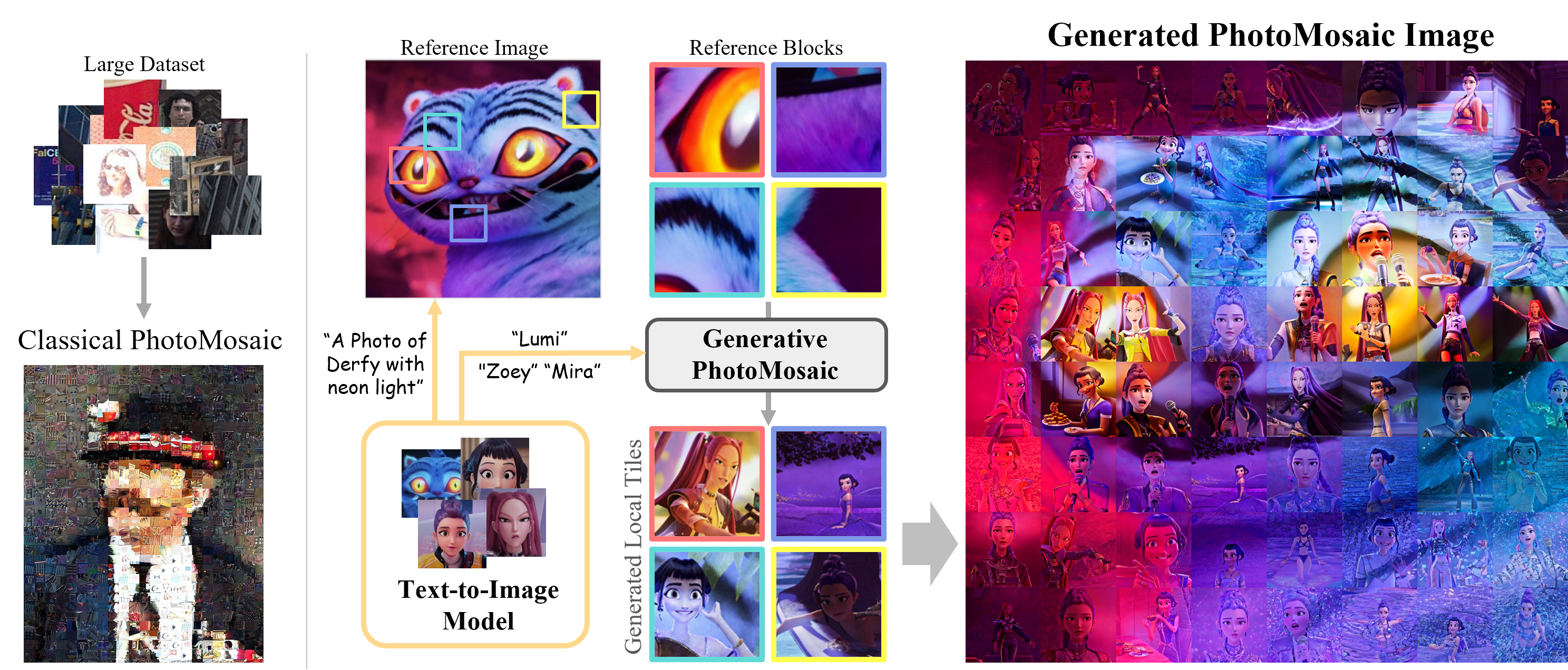}
    \vspace{-5mm}
    \captionof{figure}{\textbf{Generative Photomosaic.} We redefine photomosaic creation as a generative process. Each tile is synthesized by a diffusion model that maintains global structural alignment and reflects the local visual characteristics of the reference image.
    }
    
    \label{fig:fig1_teaser}
\end{center}

\begin{abstract}
We present the first generative approach to photomosaic creation. Traditional photomosaic methods rely on a large number of tile images and color-based matching, which limits both diversity and structural consistency. Our generative photomosaic framework synthesizes tile images using diffusion-based generation conditioned on reference images. A low-frequency–conditioned diffusion mechanism aligns global structure while preserving prompt-driven details. This generative formulation enables photomosaic composition that is both semantically expressive and structurally coherent, effectively overcoming the fundamental limitations of matching-based approaches. By leveraging few-shot personalized diffusion, our model is able to produce user-specific or stylistically consistent tiles without requiring an extensive collection of images. \href{https://robot0321.github.io/GenerativePhotomosaic/index.html}{Project page}.
\end{abstract}

\section{Introduction}
A \textit{photomosaic} is a form of image composition in which a large \textit{reference image}, also called the main image, is reconstructed from a collection of small mosaic images, referred to as \textit{tile images}. Each tile contributes local visual details, while the overall arrangement collectively forms a coherent global structure.
Owing to this dual-layered nature—global semantics emerging from local image composition—photomosaic have found applications across a wide range of domains, including industrial design, digital art, education, and advertising~\cite{silvers1997photomosaics,finkelstein1998image}. 
Depending on the semantic relationship between the reference image and its constituent tiles, photomosaic can convey diverse meanings and aesthetic impressions. 
For example, a mosaic may be created by assembling personal photographs to depict a record of past travels, by composing concert images of an artist to form the artist’s portrait, or by arranging departmental photos of a company to construct its corporate logo. 
In artistic contexts, photomosaic is also used to produce paradoxical or ironic effects by intentionally contrasting the global image with the local tile content. 
This interplay between global coherence and local diversity makes photomosaic an expressive and versatile medium for both artistic storytelling and visual communication.

Photomosaic aims to reconstruct the low-frequency structure of a reference image while allowing visual diversity in high-frequency details.
Although photomosaic has a long history as a creative image composition technique~\cite{battiato2006survey}, conventional approaches suffer from several intrinsic limitations: the difficulty of finding tiles that are structurally aligned with reference blocks from a finite image pool. 
To mitigate this, prior work adopts finer block partitioning with color adjustment.
While finer block partitioning improves low-frequency fidelity, it inherently fixes the perceptual scale of the mosaic, constraining the viewing distance at which individual tiles become discernible. This limits the ability to control how tile-level details emerge across different viewing scales.
Moreover, as finer partitions require a larger number of tiles, the limited size of the image pool often leads to repetitive use of identical tiles.
Consequently, these limitations motivate a shift beyond retrieval-based photomosaic methods toward more flexible and expressive alternatives.

To overcome these limitations, we propose a \textit{generative photomosaic} framework that leverages the expressive power of modern diffusion-based image generators. Instead of depending on large external tile datasets, our method synthesizes diverse tile images conditioned on the reference image itself, ensuring structural alignment at the global level while maintaining stylistic flexibility at the local level. Specifically, we employ a frequency-aware generation strategy that aligns low-frequency components of the generated tiles with the reference structure, while allowing high-frequency variations to be guided by text conditions. This enables both coherent reconstruction and semantically meaningful diversity within the mosaic composition.
Furthermore, to support personalized or domain-specific applications, we integrate few-shot image-conditioned diffusion models that can adapt to a small set of user-provided images. This personalization enables the photomosaic to embed personal memories, stylistic preferences, or brand-specific visual elements, thereby enriching its expressive and emotional depth. By unifying structural consistency, generative diversity, and user controllability, our framework reinterprets photomosaic creation as a generative process rather than a combinatorial one.

In summary, our main contributions are as follows:
\begin{itemize}
\item We introduce the first generative approach for photomosaic that synthesizes structurally aligned tile images without relying on large-scale external tile collections.
\item We design a structural-guided diffusion control mechanism that harmonizes low-frequency structural alignment with high-frequency detail generation, achieving both global coherence and local diversity.
\item We employ a few-shot personalized diffusion model to incorporate user-specific or domain-adaptive imagery, enhancing semantic richness and individuality.
\end{itemize}




\section{Related work}

\noindent\textbf{Photomosaic.} 
A photomosaic represents a large image reconstructed from a collection of small, uniformly sized photographic tiles.
The technique produces a coherent global perception when viewed from a distance, while each tile shows its own detail up close, making it popular in advertising, design, and digital art.
The concept of photomosaics traces back to the 1990s, popularized by Robert Silvers’ work at MIT~\cite{silvers1996photomosaics}, which first formalized the idea of reconstructing a large image using numerous smaller photographic tiles. 
Early photomosaics were primarily artistic or commercial in nature, emphasizing aesthetic composition rather than algorithmic optimization.
At that time, photomosaic was costly and time-consuming due to the large amount of computation required.
Subsequent research transformed the process into a computational task involving color histogram matching~\cite{finkelstein1998image}, tone mapping, and block-to-tile correspondence search. 
During the 2000s and 2010s, numerous heuristic and evolutionary algorithms were proposed to improve matching efficiency and reduce tile repetition~\cite{he2019composing,li2020generating}. 
These methods, however, fundamentally depended on the availability and diversity of external tile databases. 
Recent extensions explored semantic matching and hybrid approaches integrating clustering or deep features, yet all remained constrained by the need to retrieve rather than synthesize tile imagery.
In contrast, our work represents the first generative formulation of photomosaics, enabling tile synthesis through diffusion models and frequency-aware conditioning.
Furthermore, to prevent redundant tile reuse that degrades the visual quality, we leverage personalized diffusion models to generate diverse, semantically aligned tile images that correspond to the content of the reference image.

\vspace{0.5em}
\noindent\textbf{2D diffusion models for Image Generation.} 
Recent advances in generative modeling have fundamentally changed the landscape of image synthesis and composition. 
Diffusion-based models such as Stable Diffusion~\cite{rombach2022high}, Imagen~\cite{saharia2022photorealistic}, and SDXL~\cite{podell2023sdxl} demonstrate remarkable controllability over semantic and structural attributes, enabling precise manipulation of spatial layouts, textures, and visual styles. 
Several works explore \textbf{structure-aware generation}, where global layout or low-frequency components guide local detail synthesis~\cite{zhang2023adding, hertz2022prompt}. 
These techniques highlight the potential of frequency-space control and cross-scale consistency for achieving spatial coherence. 

Building on these advances, our work introduces the first \textit{generative photomosaic} framework, integrating personalized diffusion with frequency-wise conditioning to synthesize tile images that are semantically expressive, structurally aligned, and unconstrained by pre-existing datasets.

\vspace{0.5em}
\noindent\textbf{Personalized image generation.} 
Personalized image generation in diffusion models focuses on adapting text-to-image (T2I) systems to represent user-specific concepts from a few reference images while maintaining identity consistency under novel prompts.
Early methods such as Textual Inversion (TI)~\cite{gal2022image} learned new text embeddings that encodes the subject, but were limited in identity fidelity. 
DreamBooth~\cite{ruiz2023dreambooth} fine-tuned the diffusion model on subject-specific images, improving identity preservation but tended to overfit to reference examples. 
Custom Diffusion~\cite{kumari2023multi} updated only cross-attention layers, enabling scalable and efficient personalization.

Recent methods have leveraged Low-Rank Adaptation (LoRA)~\cite{hu2022lora}, a low-rank matrix decomposition technique that enables parameter-efficient fine-tuning.
DB-LoRA~\cite{ryu2023low} integrates LoRA into DreamBooth, achieving comparable fidelity to full weight fine-tuning in single-concept customization while being lightweight.
Mix-of-Show~\cite{gu2023mix} introduces ED-LoRA for single-concept tuning, which enhances expressiveness of the embedding through layer-wise embedding and multi-world representation. 
It further combines multiple LoRA modules via gradient fusion and regional sampling to enable coherent multi-concept composition.
Following recent works~\cite{po2024orthogonal, simsar2025loraclr, chen2024personalizing, kong2024omg, jiang2025mc} focused on achieving more flexible concept composition and mitigating interference among multiple concepts.

In this work, we generate personalized tile images using Mix-of-Show approach, as it provides stable and high fidelity in single-conept customization.
\begin{figure*}[t]
    \centering
    \vspace{-4mm}
    \includegraphics[width=1.0\linewidth]{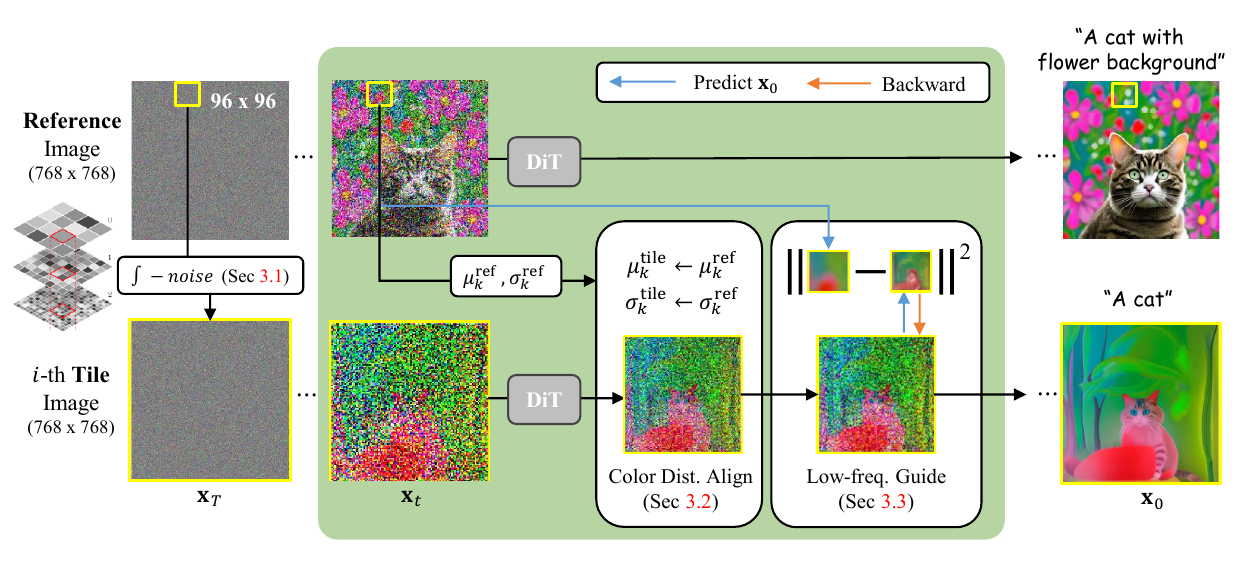}
    \vspace{-6mm}
    \captionof{figure}{\textbf{Method Overview.} We generate photomosaic images using a diffusion-based framework. During the noise initialization stage, each partial reference block (e.g. $96\times96$) is expanded to proper resolution (e.g. $768\times768$) using an integral-noise subsampling. At every denoising step, we align the color distribution of the evolving tile image with its corresponding reference block and apply a low-frequency guidance term to preserve coarse structural layout. As a result, the model produces tile images that both follow the structure of the reference image and comply with the target prompt.
    }
    \label{fig:method}
    \vspace{-2mm}
\end{figure*}

\section{Method}
We aim to construct a photomosaic image $I^{\text{mosaic}}$ that represents a given reference image $I^{\text{ref}}$ through a set of tile images $\{ I^{\text{tile}}_k \}$. The reference image $I^{\text{ref}}$ is partitioned into non-overlapping patches $\{ P^{\text{ref}}_k \}$, and our goal is to preserve the color statistics and coarse structures of these patches while injecting target-prompt–aligned textures or fine details into each local tile.

We propose several techniques that guide the generation of tile images $\{ I^{\text{tile}}_k \}$ such that both the global reference patches $\{ P^{\text{ref}}_k \}$ and the target text prompt are simultaneously satisfied. Section~\ref{sec:method_init} presents a noise initialization strategy designed to improve consistency across tiles. Section~\ref{sec:method_adain} introduces latent distribution matching that aligns color distributions. Section~\ref{sec:method_sync} describes a gradient-descent–based latent adjustment that enforces coarse structural guidance during generation. 



\subsection{Integral-Noise Subsampling for Initialization} \label{sec:method_init}
To obtain stable and spatially consistent tile-level latents, we generate fine-scale Gaussian noise conditioned on the global block rather than relying on independently sampled random initialization. Random latents injected per tile often introduce incoherent high-frequency patterns and step-to-step instability during denoising. Instead, we construct a coherent multi-scale noise field by adapting the integral-noise subsampling~\cite{chang2024how}, which interprets each pixel as the integral of an underlying continuous white Gaussian field.

Given a global latent $B_k \in \mathbb{R}^{h \times w}$ and a scale factor $s$ that maps each global block to $s^2$ fine pixels (e.g., $96\!\rightarrow\!768$, so $s=8$), the conditional distribution of the fine-scale latent over a block is
\begin{equation}
W \mid X
\;\sim\;
\frac{X}{N}\mathbf{u}
\;+\;
\frac{1}{\sqrt{N}}
\big(Z - \langle Z\rangle\mathbf{u}\big),
\label{eq:integral-noise}
\end{equation}
where $X$ denotes a coarse block-level noise variable, and $W$ is the corresponding fine-scale noise conditioned to preserve the block integral. Here $Z\sim\mathcal{N}(0,I)$ is Gaussian distribution,  $N=s^2$, $\mathbf{u}$ is an all-ones vector of length $N$. This decomposition ensures that the fine-scale latent exactly preserves the coarse integral while the zero-sum residual recovers the correct conditional covariance of a Gaussian field. As a result, all tiles originate from a single coherent noise source, maintaining structural consistency across regions and avoiding texture discontinuities that commonly arise with random per-tile initialization.

Diffusion models, however, assume a unit-variance per-pixel Gaussian prior, whereas the conditional form in Eq.~\ref{eq:integral-noise} yields marginal variance $(N{-}1)/N < 1$. To align the subsampled latent with the diffusion prior, we apply a deterministic variance normalization in the zero-sum subspace:
\begin{equation}
\widetilde{W} = \frac{X}{N}\mathbf{u}\;+\;\frac{1}{\sqrt{1 - \tfrac{1}{N}}} \Big(W - \tfrac{X}{N}\mathbf{u}\Big).
\label{eq:variance-normalization}
\end{equation}
where $\widetilde{W}$ is the variance-normalized noise that matches the unit-variance diffusion prior while maintaining block-level coherence, following the presentations in ~\cite{chang2024how}. 
This transformation preserves the coarse integral exactly while restoring unit variance at every fine-scale pixel, yielding a \emph{diffusion-ready} coherent noise prior that is statistically compatible with the pretrained model.

In practice, initializing tiles with this coherent multi-scale noise leads to significantly more stable denoising dynamics. During early timesteps, where the model is highly sensitive to high-frequency perturbations, the integral-noise initialization reduces variance spikes, prevents boundary artifacts between tiles, and produces globally consistent structure compared to independent random noise sampling.

\subsection{Color Distribution Alignment through Adaptive Instance Normalization} \label{sec:method_adain}
To ensure coherent color appearance across tiles, we adopt a lightweight
distribution matching strategy inspired by adaptive instance normalization (AdaIN)
used in style transfer. While each tile in our framework is generated independently,
direct composition without color normalization often yields noticeable tone
inconsistencies and undesired color seams across tile boundaries.

For each tile $I_k^{\text{tile}}$ and the corresponding reference block $B_k$ from the global
guidance image, we compute their first-order and second-order statistics:
\begin{equation}
\mu_k^t, \sigma_k^t = \text{Mean}(I_k^{\text{tile}}),\; \text{Std}(I_k^{\text{tile}}),
\end{equation}
\vspace{-1em}
\begin{equation}
\mu_k^r, \sigma_k^r = \text{Mean}(B_k),\; \text{Std}(B_k).
\end{equation}
We then adjust each tile to match the global color distribution:
\begin{equation}
\hat{I}_k^{\text{tile}} = 
\mu_k^r + \sigma_k^r \odot
\frac{I_k^{\text{tile}} - \mu_k^t}{\sigma_k^t}.
\end{equation}

This simple normalization aligns the per-tile color statistics with the global
target, effectively harmonizing color tones across the entire mosaic.
Despite its simplicity, we find that this distribution matching significantly
reduces color drifting artifacts and improves global visual consistency.

\subsection{Low-Frequency Structural Guidance with Gradient Descent} \label{sec:method_sync}

To ensure global structural consistency across tiles while still allowing local appearance diversity, we incorporate a tile-wise low-frequency guidance step into the denoising process. At each diffusion timestep $t$, given a local latent $z_k^t$ for tile $k$, we first predict the corresponding clean latent $\hat{x}_{0,i}(t)$ using the DDIM $x_0$ prediction formulation:
\begin{equation}
\hat{x}_{0,k}(t) = \alpha_t\, z_k^t - \sigma_t\, \varepsilon_\theta(z_k^t, t, c),
\end{equation}
where $\alpha_t = \sqrt{\bar{\alpha}_t}$ and $\sigma_t = \sqrt{1-\bar{\alpha}_t}$ follow the DDPM schedule, and $\varepsilon_\theta$ denotes the noise prediction network conditioned on $c$.
We decode $\hat{x}_{0,k}(t)$ using the VAE decoder and resize the resulting RGB patch to the tile resolution. Letting $\mathcal{G}_\sigma$ denote a fixed low-pass operator (e.g., Gaussian blur), we enforce consistency between the low-frequency components of the generated tile and those of the global guiding tile $\tilde{y}_i$:
\begin{equation}
\ell_k(t) = \left\|\mathcal{G}_\sigma\!\big(I^{tile}_k(t)\big) - \mathcal{G}_\sigma(B_k(t)) \right\|_2^2.
\end{equation}

We then directly update the latent $z_i^t$ via gradient descent on $\ell_i(t)$:
\begin{equation}
z_k^t \leftarrow z_k^t - w \, \nabla_{z_k^t} \ell_k(t),
\qquad
w \leftarrow \gamma w,
\end{equation}
where $w$ is a per-tile update magnitude and $\gamma<1$ (e.g., $\gamma = 0.95$) applies exponential decay over diffusion steps. Importantly, unlike SyncDiffusion, which aligns perceptual similarity via LPIPS at the image level, our objective focuses solely on low-frequency structural alignment. This encourages global coherence and spatial layout consistency while leaving high-frequency details free to adapt to local content and style variations.

\section{Experiments} 


\subsection{Comparison Methods}
We compare our generative photomosaic framework with both traditional and diffusion-based baselines.
As classical non-generative methods, we adopt feature matching and tone adjustment (Match \& Tone) ~\cite{lee2014generation}. This approach replaces each image block with the most similar tile in terms of L2 distance within a given image pool, following the conventional photomosaic paradigm. The tone mapping method modifies the tone of a reference or arbitrary image to match each target block, enabling photomosaic construction even with a single source image. In contrast, histogram matching aligns the pixel intensity distribution of a reference image to each block, achieving more accurate tone transfer but still lacking spatial structure preservation.

As we introduce the first generative approach to photomosaic creation, we explored and adapted a range of existing ideas and techniques that could be repurposed for tile generation within the photomosaic pipeline, including Color ControlNet, AdaIN, NoiseBlend, and StreamDiffusion I2I.
Color ControlNet (Stable Diffusion v1.4~\cite{Rombach_2022_CVPR} with T2Iadapter~\cite{mou2024t2i}) uses both a low-resolution reference and a text prompt as conditions, aiming to maintain color coherence while adhering to textual semantics. AdaIN (Adaptive Instance Normalization)~\cite{huang2017arbitrary} re-normalizes the mean and variance of denoised features at each step to match those of the corresponding global block, facilitating color tone consistency across patches. NoiseBlend~\cite{lee2024diffusion} blends denoised results from the global and local branches at every diffusion step in fixed ratios, seeking to preserve global structure while injecting local texture variations. Finally, StreamDiffusion Image-to-Image (I2I)~\cite{kodaira2025streamdiffusion} refines each low-resolution tile via the image-to-image mode of StreamDiffusion under text guidance, providing an alternative diffusion-based generation path focused on local refinement.

\begin{figure*}[t]
    \centering
    \includegraphics[width=1\linewidth]{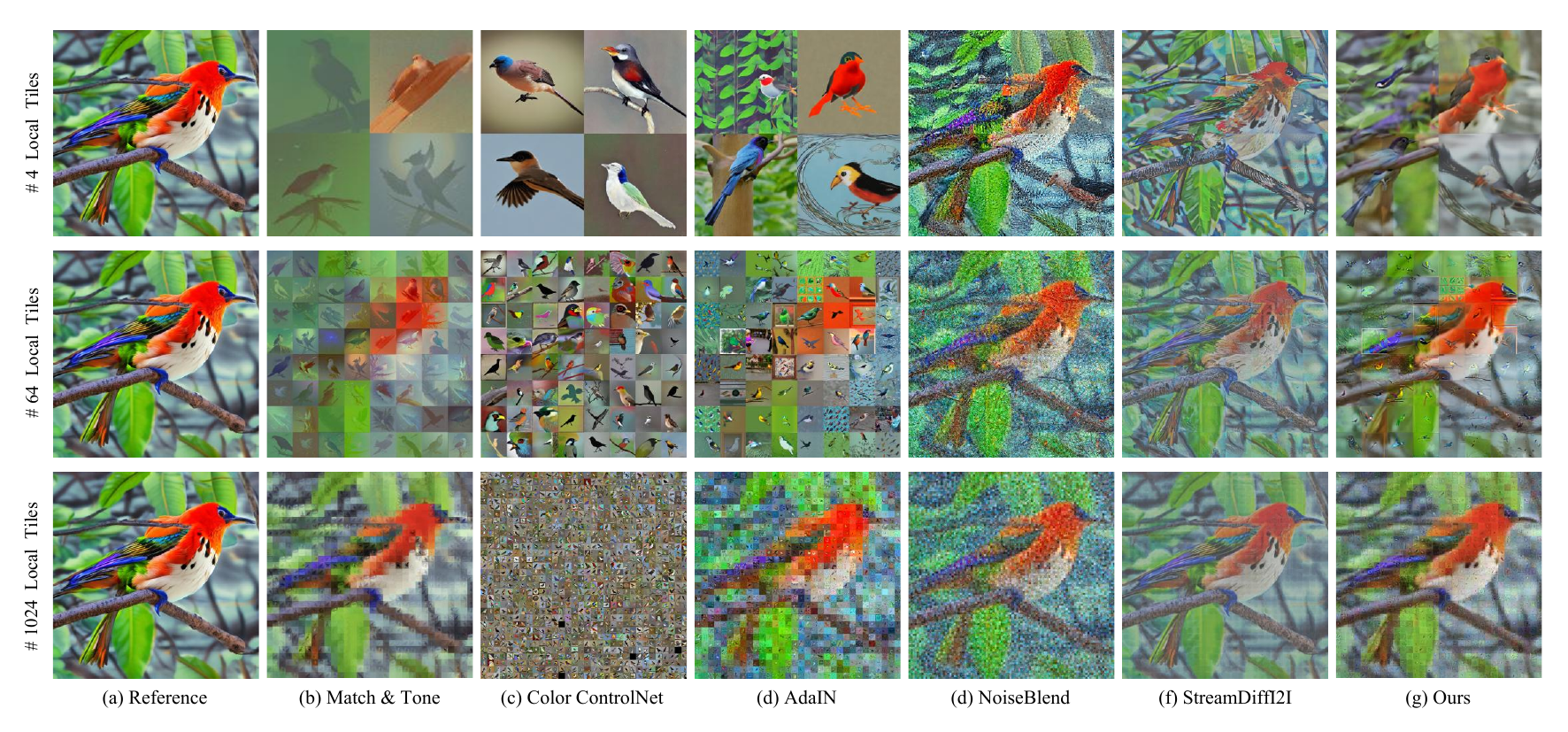}
    \vspace{-8mm}
    \captionof{figure}{\textbf{Qualitative Results across Different Mosaic Levels.}}
    \label{fig:comparison}
    \vspace{-4mm}
\end{figure*}

\subsection{Experimental Setup}
\noindent\textbf{Implementation Details.} 
We build our generative photomosaic framework on Stable Diffusion 2.1, generating all images at $768 \times 768$ resolution. 
We use 50 inference steps with guidance scale of 7.5. 
For photomosaic construction, we divide the image into $2^L \times 2^L$ tiles, where $L$ is the mosaic level (e.g., $L = 3$ yields an $8 \times 8$ mosaic). 
$w$ is fixed to 5000 for quantitative evaluation. 
All experiments are conducted on a single NVIDIA A6000 GPU, and generating one photomosaic image takes about 15 minutes.
For personalized photomosaic generation, we tuned a separate LoRA module for each concept, injecting it into the linear layers of all attention modules in both Unet and text encoder, using a rank of $r=4$, following the configuration used in Mix-of-Show~\cite{gu2023mix}.

\begin{table}[t]
    \centering
    \caption{\textbf{Quantitative results}}
    \resizebox{\linewidth}{!}{
    \begin{tabular}{c|cccc>{\centering\arraybackslash}p{0.9cm}|ccccc}
        \toprule
        & \multicolumn{5}{c|}{Ref. Image (Global)} & \multicolumn{5}{c}{Tile Image (Local)} \\
        & PSNR$_{\uparrow}$ & SSIM$_{\uparrow}$ & LPIPS$_{\downarrow}$ &  HPSv2$_{\uparrow}$ & IR$_{\uparrow}$  & BLIP$_{\uparrow}$ & CLIP$_{\uparrow}$ & IQA$_{\uparrow}$ &  HPSv2$_{\uparrow}$ & IR$_{\uparrow}$  \\
        \midrule
        Match \& Tone~\cite{lee2014generation} & 16.719  & 0.366 & 0.206 & 0.129 & -1.783  & 0.625 & 0.612 & 0.596 & 0.174 & -1.145 \\
        Color ControlNet~\cite{mou2024t2i} & 10.145 & 0.080 & 0.435 & 0.123 & -2.272 & 0.639 & 0.613 & 0.811 & 0.266 & 0.472  \\
        AdaIN~\cite{huang2017arbitrary} & 12.983 & 0.204 & 0.400 & 0.130 & -2.279  & 0.577 & 0.614 & 0.834 & 0.244 & -0.106 \\
        NoiseBlend~\cite{lee2024diffusion} & 11.590 & 0.205 & 0.479 & 0.117 & -2.275 & 0.239 & 0.597 & 0.385  & 0.195 & -1.851  \\
        StreamDiffI2I~\cite{kodaira2025streamdiffusion} & 15.212 & 0.414 & 0.272 & 0.127 & -1.377 & 0.166 & 0.602 & 0.259 & 0.173 & -1.073 \\
        \midrule
        Ours & 17.598 & 0.551 & 0.121 &0.132 & -2.225 & 0.569 & 0.619 & 0.812 & 0.239 & -0.317 \\
        \bottomrule
    \end{tabular}
    }
    \label{tab:quan_result}
\end{table}

\begin{figure}[t]
\centering

\begin{minipage}[t]{0.36\linewidth}
  \vspace{0pt}
  \centering
  \captionsetup{type=table}
  \caption{\footnotesize \textbf{User Study}}
  \label{tab:user}
  \vspace{1mm}

  \resizebox{0.95\linewidth}{!}{%
    \begin{tabular}{lcc}
    \toprule
    & \multicolumn{2}{c}{Ours Win Rate (\%)} \\
    & Global & Local \\
    \midrule
    Match\&Tone        & 98.7 & 73.3 \\
    Color ControlNet   & 95.7 & 7.3 \\
    AdaIN              & 96.7 & 12.7 \\
    NoiseBlend         & 15.0 & 92.3 \\
    StreamDiffI2I      & 23.0 & 88.7 \\
    \bottomrule
\end{tabular}
  }
\end{minipage}
\hfill
\begin{minipage}[t]{0.62\linewidth}
  \vspace{0pt}
  \centering
  \input{fig/comparison_plot}
  \vspace{-2mm}
  \caption{\textbf{Comparison of Quantitative results.}}
  \label{fig:comparison_plot}
\end{minipage}

\vspace{-2mm}
\end{figure}

\begin{figure*}
    \centering
    \includegraphics[width=1.0\linewidth]{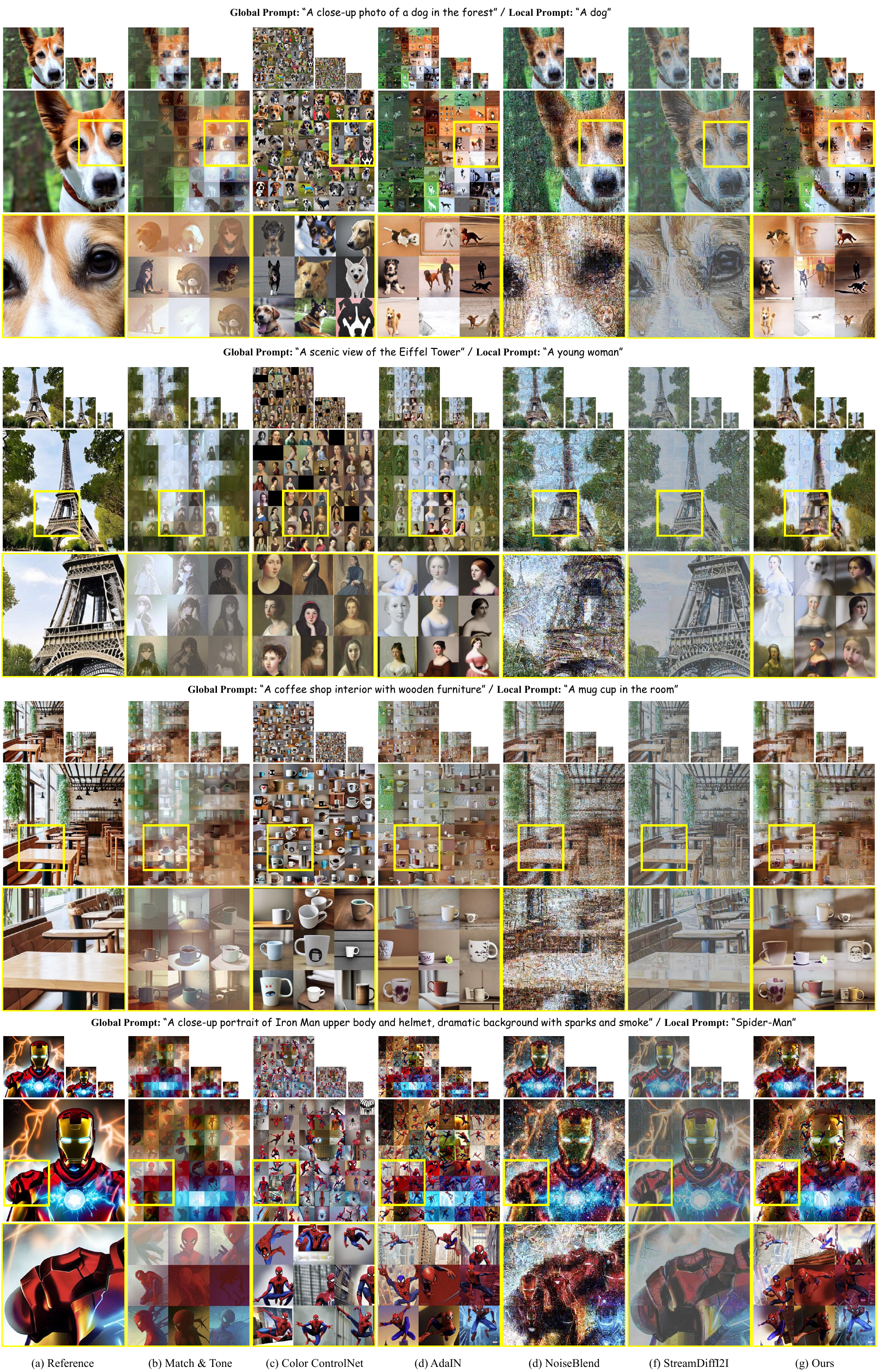}
    \vspace{-6mm}
    \captionof{figure}{\textbf{Qualitative Results of Generative Photomosaic.}}
    \label{fig:main_comparison}
    \vspace{-2mm}
\end{figure*}

\vspace{0.5em}
\noindent\textbf{Metric.} 
We evaluate photomosaic quality using metrics that capture both global structural fidelity and local tile realism. 
PSNR and SSIM~\cite{wangzhou2004image} measure low and mid frequency reconstruction quality, while LPIPS~\cite{zhang2018unreasonable} evaluates perceptual structural similarity. 
These metrics reflect how well the global structure of the reference image is preserved, where the reference images generated by Stable Diffusion are used as ground truth.
Also, BLIP~\cite{li2022blip} and CLIP~\cite{radford2021learning} scores evaluate semantic alignment of the local tiles with their corresponding prompts, and CLIP-IQA~\cite{wang2023exploring} measures perceptual tile quality, for which we report the score of the "quality" attribute.
We additionally evaluated metrics designed to align with human preference, thereby better capturing perceptual quality, such as Human Preference Score v2 (HPSv2)~\cite{wu2023human} and Image Reward (IR)~\cite{xu2023imagereward}.

\vspace{0.5em}
\noindent\textbf{Evaluation Details.} 
We utilize 12 global and local prompt sets and evaluate all methods with 15 fixed random seeds for reproducibility, ensuring comparable quantitative metrics across methods. 
The evaluation prompts to generate photomosaic images are presented in supplementary material.
For quantitative evaluation, we measure metrics for global structural fidelity across multiple resolutions (32, 64, 128, 256). PSNR, SSIM, and LPIPS are reported at resolution 64 for the HPSv2 and IR metrics are reported at resolution 128.

\subsection{Evaluation}
Figure~\ref{fig:comparison} compares baseline methods derived from prior work across different mosaic levels. \Cref{tab:quan_result} and Figure~\ref{fig:comparison_plot} reports the quantitative comparisons. Our method maintains strong performance across both global and local metrics, without being biased toward either aspect.
Figure~\ref{fig:main_comparison} presents generative mosaics created with various prompts and evaluates their performance against existing baselines. Our method preserves the global structure more faithfully while leveraging the reference block layout and generating tiles that follow the local prompts.

Traditional methods such as feature matching, tone adjustment, and histogram matching fail to maintain structural consistency between global and local regions. Because these approaches depend solely on local tone or color similarity, they require either a large number of tiles or very small patch sizes to preserve the recognizable global layout. As a result, both image collection and generation incur high computational and data costs.

The diffusion-based methods demonstrate distinct trade-offs between structural coherence, texture realism, and prompt alignment. Color ControlNet, although conditioned on a low-resolution reference, fails to adequately reflect its structural guidance. Regardless of the guidance strength, the model tends to prioritize text alignment, resulting in over-saturated or misaligned compositions. AdaIN successfully transfers the overall color tone of each global block but lacks geometric precision, making it more suitable for fine-grained mosaics rather than globally structured ones. NoiseBlend effectively preserves the structural layout of the global image while injecting detailed local textures. However, blending denoised predictions at each step introduces noisy artifacts, which appear to stem from conflicting denoising directions guided by the global and local prompts. StreamDiffusion (I2I) maintains low-frequency consistency and generates coherent local textures but fails to fully reflect the intended prompt semantics, yielding visually plausible yet semantically under-constrained results.

\begin{table}[t]

    \setlength{\tabcolsep}{5pt}
    \centering
    \caption{\textbf{Ablation Results.}}
    \vspace{-2mm}
    \resizebox{0.8\linewidth}{!}{
    \begin{tabular}{c|ccc|ccc}
        \toprule
        & \multicolumn{3}{c|}{Ref. Image (Global)} & \multicolumn{3}{c}{Tile Image (Local)} \\
        & PSNR$_{\uparrow}$ & SSIM$_{\uparrow}$ & LPIPS$_{\downarrow}$ & BLIP$_{\uparrow}$ & CLIP$_{\uparrow}$ & IQA$_{\uparrow}$  \\
        \midrule
         \textbf{Ours} & \textbf{17.598} & \textbf{0.551} & \textbf{0.121} & \textbf{0.569} & \textbf{0.619} & \textbf{0.812} \\
        w/o sec 3.1 & 16.224&0.510&0.186&0.562&0.611&0.806 \\
        w/o sec 3.2 & 16.431&0.495&0.161&0.566&0.612&0.564 \\
        w/o sec 3.3 & 13.142&0.314&0.398&0.567&0.614&0.810 \\
        \bottomrule
    \end{tabular}
    }
    \vspace{-4mm}
    \label{tab:seg}
\end{table}
\begin{figure}[t]
    \newcommand{\ww}{0.24\linewidth}
    \centering

    \begin{subfigure}{\ww}
        \centering
        \includegraphics[width=\linewidth]{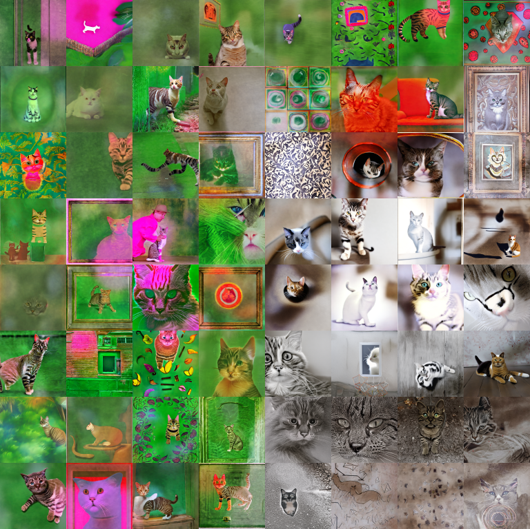} 
        \caption{Ours}
        \label{fig:ablation_ours}
    \end{subfigure}
    \hfill
    \begin{subfigure}{\ww}
        \centering
        \includegraphics[width=\linewidth]{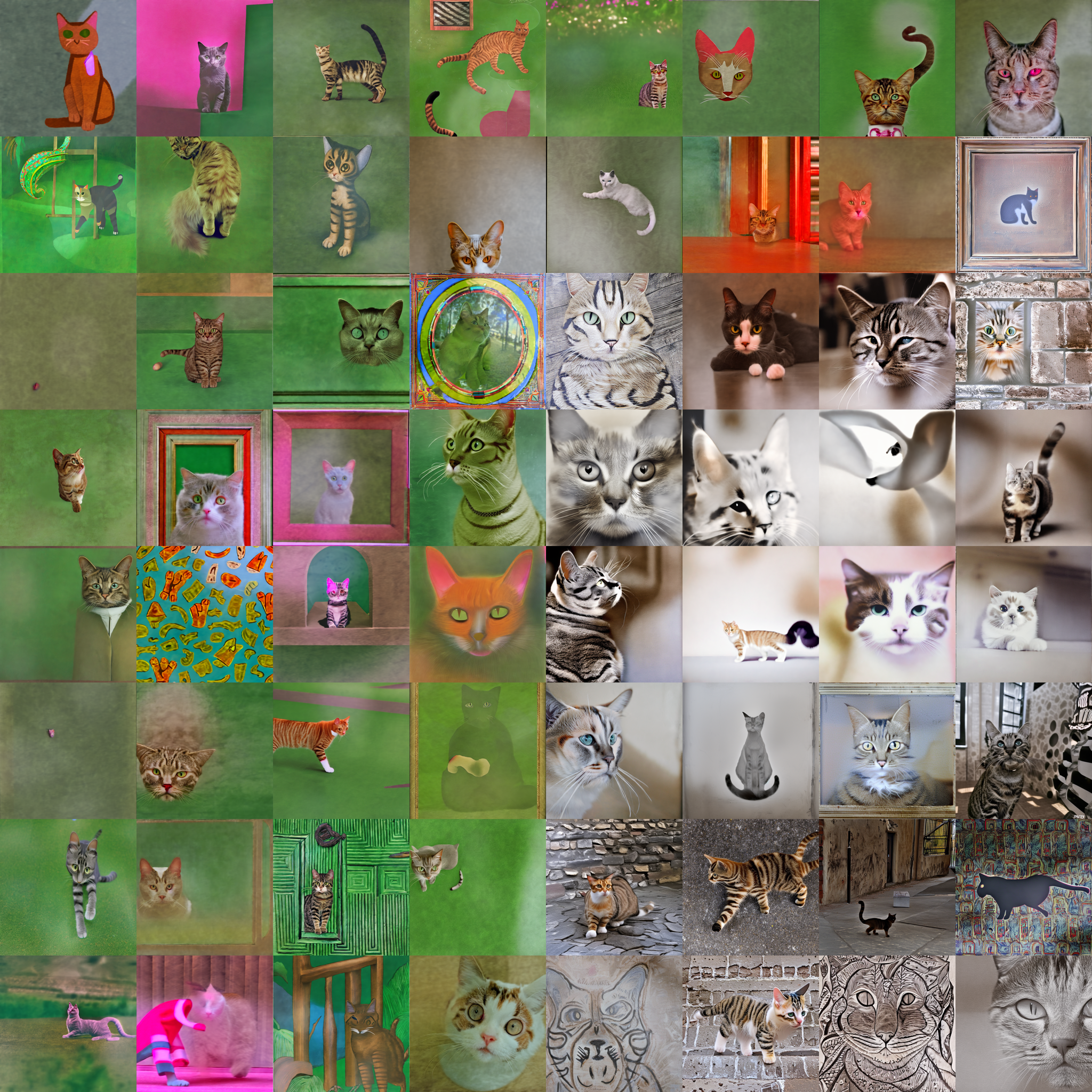}
        \caption{w/o $\int$-noise Init}
        \label{fig:ablation_init}
    \end{subfigure}
    \hfill
    \begin{subfigure}{\ww}
        \centering
        \includegraphics[width=\linewidth]{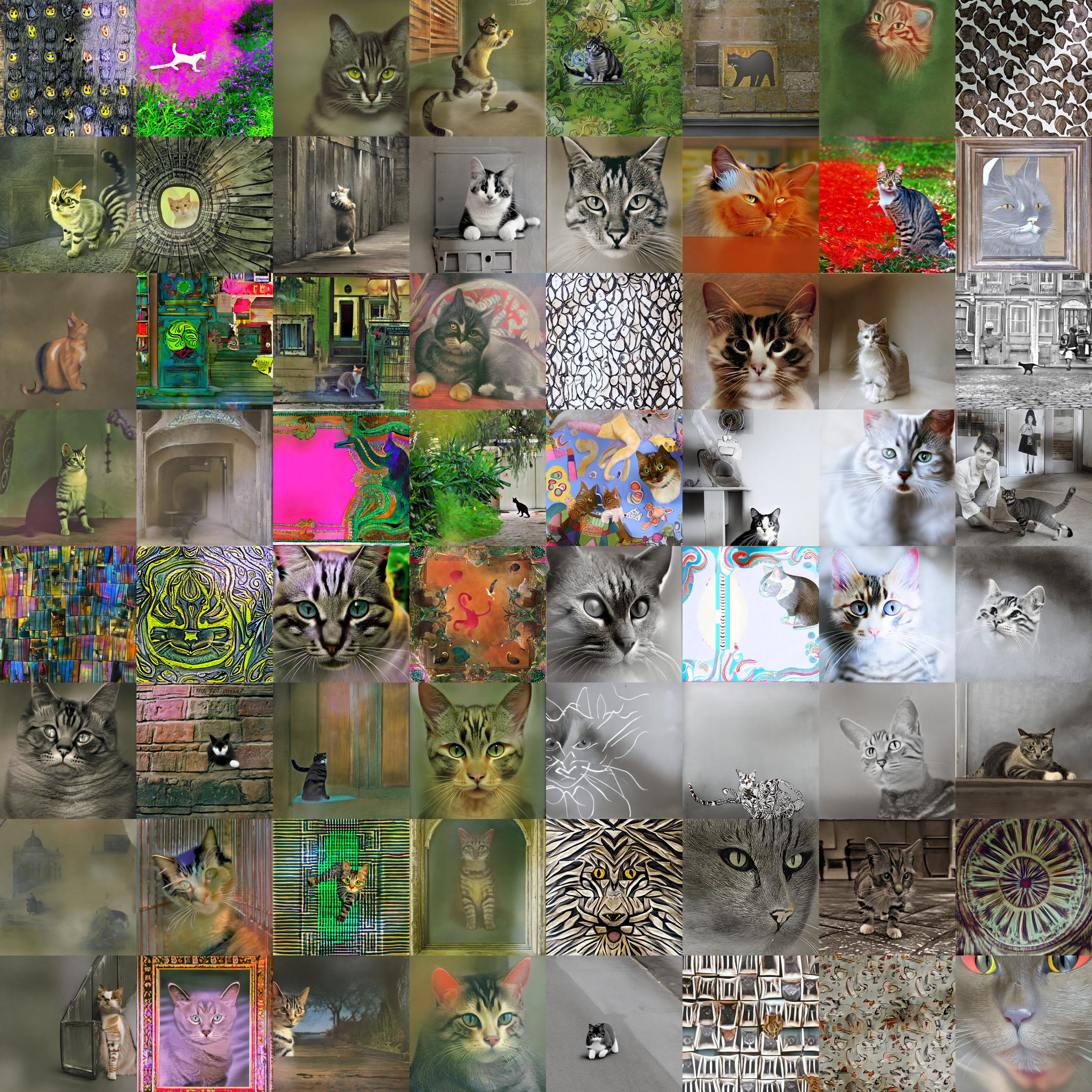}
        \caption{w/o Color Align}
        \label{fig:ablation_adain}
    \end{subfigure}
    \hfill
    \begin{subfigure}{\ww}
        \centering
        \includegraphics[width=\linewidth]{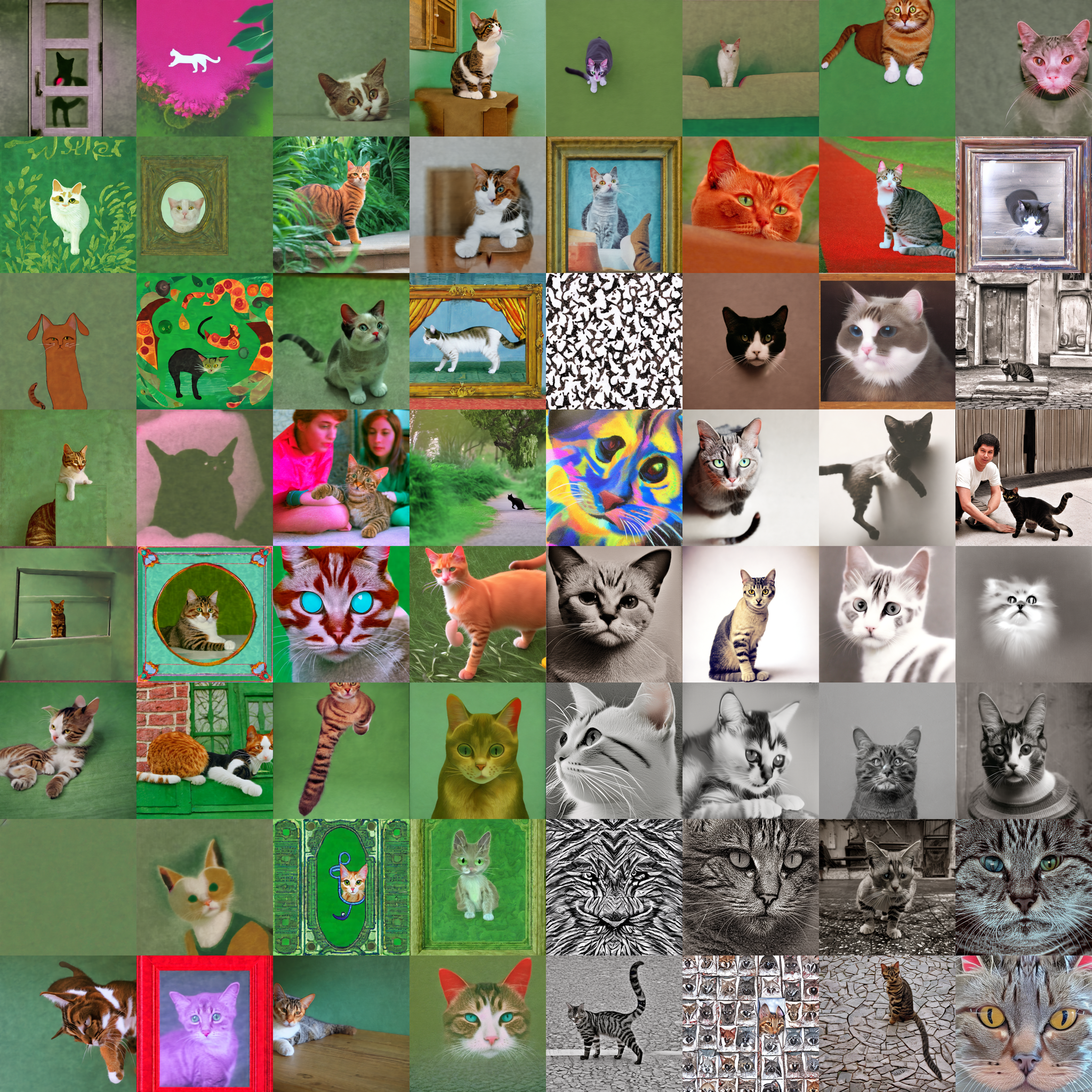}
        \caption{w/o Gradient Guide}
        \label{fig:ablation_sync}
    \end{subfigure}

    \vspace{-2mm}
    \caption{\textbf{Visualization of Ablation Results.}}
    \label{fig:ablation}
    \vspace{-4mm}
\end{figure}

\vspace{0.5em}
\noindent\textbf{Human Preference Study.} 
We conducted an A/B preference test with 100 participants, comparing our method against each baseline. 
For each of the five baselines, three comparison examples were included, resulting in a total of 15 questions. 
The \Cref{tab:user} reports the percentage of responses in which our method was preferred over the baseline.
Human A/B test results reveal a clear global–local trade-off in baselines.
Baselines that achieve high global fidelity exhibit limited local tile quality, while locally emphasized methods degrade global semantic structure.
Our method maintains a balanced performance across both aspects.

\begin{table}[t]
    \vspace{-1mm}
    \setlength{\tabcolsep}{5pt}
    \centering
    \caption{\textbf{Guidance Ablation Results.}}
    \vspace{-2mm}
    
    \resizebox{0.8\linewidth}{!}{
    \begin{tabular}{c|ccc|ccc}
        \toprule
        & \multicolumn{3}{c|}{Ref. Image (Global)} & \multicolumn{3}{c}{Tile Image (Local)} \\
        & PSNR$_{\uparrow}$ & SSIM$_{\uparrow}$ & LPIPS$_{\downarrow}$ & BLIP$_{\uparrow}$ & CLIP$_{\uparrow}$ & IQA$_{\uparrow}$ \\
        \midrule
         \textbf{Ours} & \textbf{17.598} & \textbf{0.551} & \textbf{0.121} & \textbf{0.569} & \textbf{0.619} & \textbf{0.812} \\
        LPIPS-guided &12.242&0.216&0.456 &0.373&0.603&0.656\\
        YCbCr-guided &14.107&0.361&0.337 &0.548&0.611&0.801\\
        \bottomrule
    \end{tabular}
    }
    \vspace{-1mm}
    \label{tab:s_ab}
\end{table}
\begin{figure*}[t]
    \centering
    \newcommand{\w}{0.24\linewidth} 

    \begin{subfigure}{\w}
        \centering
        \includegraphics[width=\linewidth]{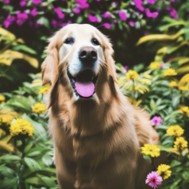}
        \caption{Reference}
    \end{subfigure}
    \hfill
    \begin{subfigure}{\w}
        \centering
        \includegraphics[width=\linewidth]{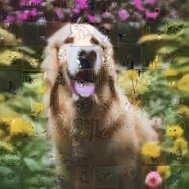}
        \caption{Ours}
    \end{subfigure}
    \hfill
    \begin{subfigure}{\w}
        \centering
        \includegraphics[width=\linewidth]{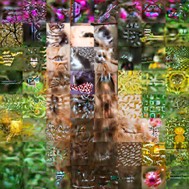}
        \caption{LPIPS-guided}
    \end{subfigure}
    \hfill
    \begin{subfigure}{\w}
        \centering
        \includegraphics[width=\linewidth]{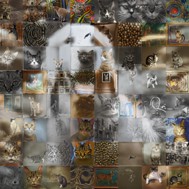}
        \caption{YCbCr-guided}
    \end{subfigure}

    \caption{
        \textbf{Visualization of Guidance Ablation Results.}
    }
    \vspace{-4mm}
    \label{fig:s_ablation}
\end{figure*}

\subsection{Ablation Study}
\vspace{0.5em}
\noindent\textbf{Component Ablation.} 
We conducted an ablation study to evaluate the contribution of each proposed component, as illustrated in \Cref{fig:ablation}.
\Cref{fig:ablation_init} shows the result obtained by replacing our integral noise subsampling with random initialization. The random initialization disrupts the latent distribution, leading to poor preservation of the structural layout of the reference image.
\Cref{fig:ablation_adain} presents the result without the proposed Color Alignment module. In this case, most color information vanishes and the output appears nearly monochromatic. This observation suggests that our gradient guidance not only constrains spatial structures but also implicitly enforces low-frequency consistency in the color space.
In \Cref{fig:ablation_sync}, the reference structure is almost entirely lost. 
%

\begin{figure*}[t]
    \centering
    \includegraphics[width=1.0\linewidth]{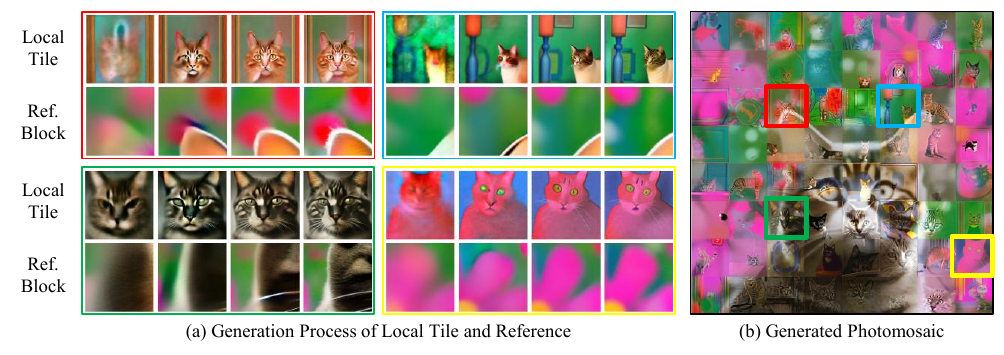}
    \vspace{-8mm} 
    \captionof{figure}{\textbf{Generation process of local tile and its corresponding reference block.}}
    \label{fig:ddiminversion}

\end{figure*}
\begin{figure*}[t]
    \centering
    \newcommand{\w}{0.16\linewidth}  

    \begin{subfigure}{\w}
        \centering
        \includegraphics[width=\linewidth]{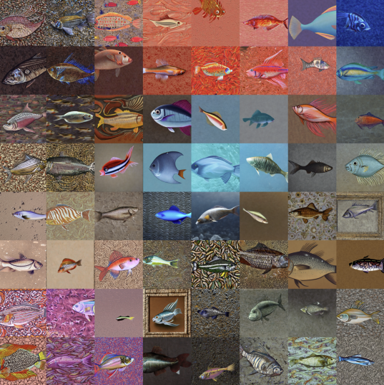}
        \caption{$w=0$}
        \label{fig:ex_a}
    \end{subfigure}
    \hfill
    \begin{subfigure}{\w}
        \centering
        \includegraphics[width=\linewidth]{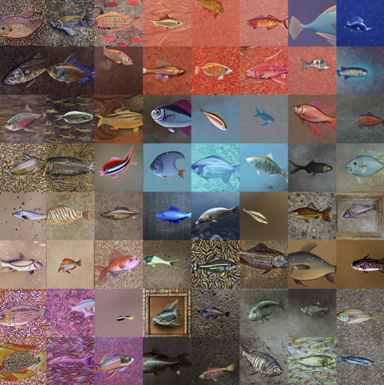}
        \caption{$w=500$}
        \label{fig:ex_b}
    \end{subfigure}
    \hfill
    \begin{subfigure}{\w}
        \centering
        \includegraphics[width=\linewidth]{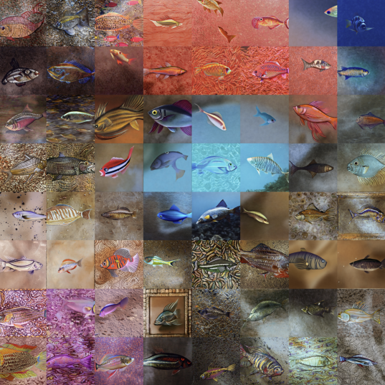}
        \caption{$w=2000$}
        \label{fig:ex_c}
    \end{subfigure}
    \hfill
    \begin{subfigure}{\w}
        \centering
        \includegraphics[width=\linewidth]{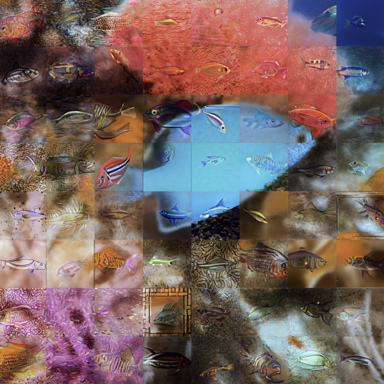}
        \caption{$w=5000$}
        \label{fig:ex_d}
    \end{subfigure}
    \hfill
    \begin{subfigure}{\w}
        \centering
        \includegraphics[width=\linewidth]{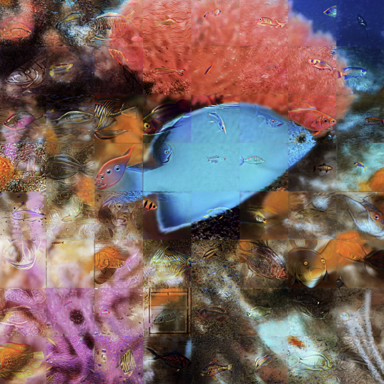}
        \caption{$w=10000$}
        \label{fig:ex_e}
    \end{subfigure}
    \hfill
    \begin{subfigure}{\w}
        \centering
        \includegraphics[width=\linewidth]{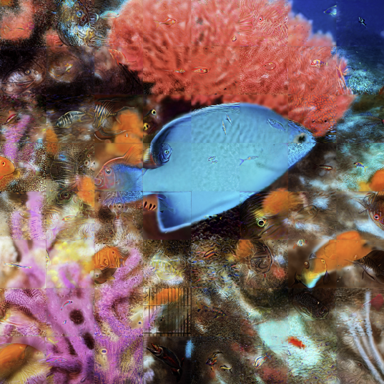}
        \caption{$w=20000$}
        \label{fig:ex_f}
    \end{subfigure}
    \caption{\textbf{Ablation Results on the Guidance Weight Magnitude.}}
    \vspace{-3mm}
    \label{fig:w_ablation}
\end{figure*}

\vspace{0.5em}
\noindent\textbf{Analysis on the Guidance Strategy.} 
Our approach leverages tile-wise low-frequency guidance during denoising process to maintain global structural consistency across tiles.
For comparison, we conducted additional experiments using alternative guidance strategies, as shown in \Cref{fig:s_ablation} and \Cref{tab:s_ab}.
First, our MSE-based low-frequency structural guidance in is inspired by SyncDiffusion~\cite{lee2023syncdiffusion}, which instead employs an LPIPS loss to enforce perceptual consistency. In practice, however, we observe that LPIPS tends to introduce undesirable artifacts rather than encouraging adherence to low-frequency structure, whereas our MSE formulation provides a more stable and structurally faithful guidance signal.
Second, we further experimented with applying the structural guidance only on the luminance. Specifically, we converted the RGB outputs into the YCbCr color space and imposed the MSE loss exclusively on the luminance channel $Y$:
\vspace{-1.2em}
\begin{align}
Y &= 0.299\,R + 0.587\,G + 0.114\,B, \\
\mathcal{L}_{Y} &= \| Y^{tile}_k - Y^{ref.block}_k \|_2^2 .
\end{align}
The motivation was that enforcing the loss on luminance would emphasize low-frequency structural alignment while leaving chrominance relatively unconstrained. However, this strategy did not yield noticeable improvement in color preservation; the resulting images still exhibited significant desaturation and chroma collapse.

\vspace{0.5em}
\noindent\textbf{Denoising Process Visualization.} 
\Cref{fig:ddiminversion} illustrates the evolution of the tile image and its corresponding reference block throughout the diffusion process. As denoising progresses, the tile image is guided to gradually acquire the structural characteristics of the reference block, enabling the effective synthesis of photomosaic.

\vspace{0.5em}
\noindent\textbf{Analysis on the Guidance Weight Magnitude.} 
To analyze the effect of the low-frequency structural guidance, we conducted an ablation experiment by varying the guidance weight $w$ used in Eq. (8), as shown in  \Cref{fig:w_ablation}. 
As $w$ controls the magnitude of the gradient-based adjustment applied to each tile latent, it determines the strength of structural alignment enforced during denoising. 
When $w$ is small, the guidance imposes only a mild constraint, preserving fine-grained visual details driven by the text prompt in local tiles.
Larger values of $w$ strengthen the low-frequency alignment, guiding each tile toward the structure of the reference.
This improves the clarity of the global structure but diminishes local semantic detail, often producing overly smoothed textures within the tiles.
These observations verify that $w$ governs the trade-off between global structural preservation and local detail fidelity.

\begin{figure*}[t]
    \newcommand{\ww}{0.32\linewidth}
    \centering

    \begin{subfigure}{\ww}
        \centering
        \includegraphics[width=\linewidth, height=2\linewidth, keepaspectratio]{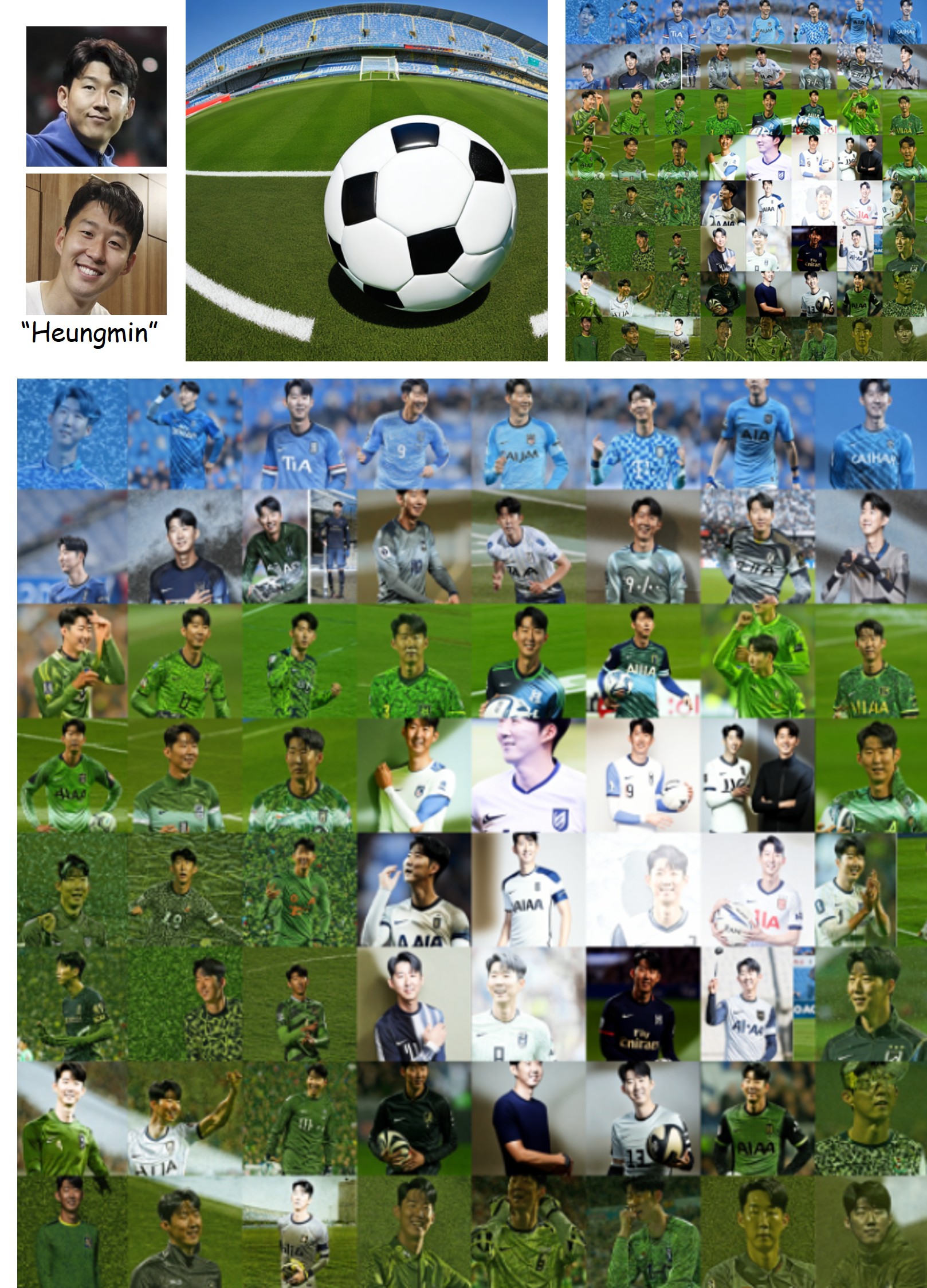}
        \caption{\centering {Personalization Example 1}: "Heungmin"}
    \end{subfigure}
    \hfill
    \begin{subfigure}{\ww}
        \centering
        \includegraphics[width=\linewidth, height=2\linewidth, keepaspectratio]{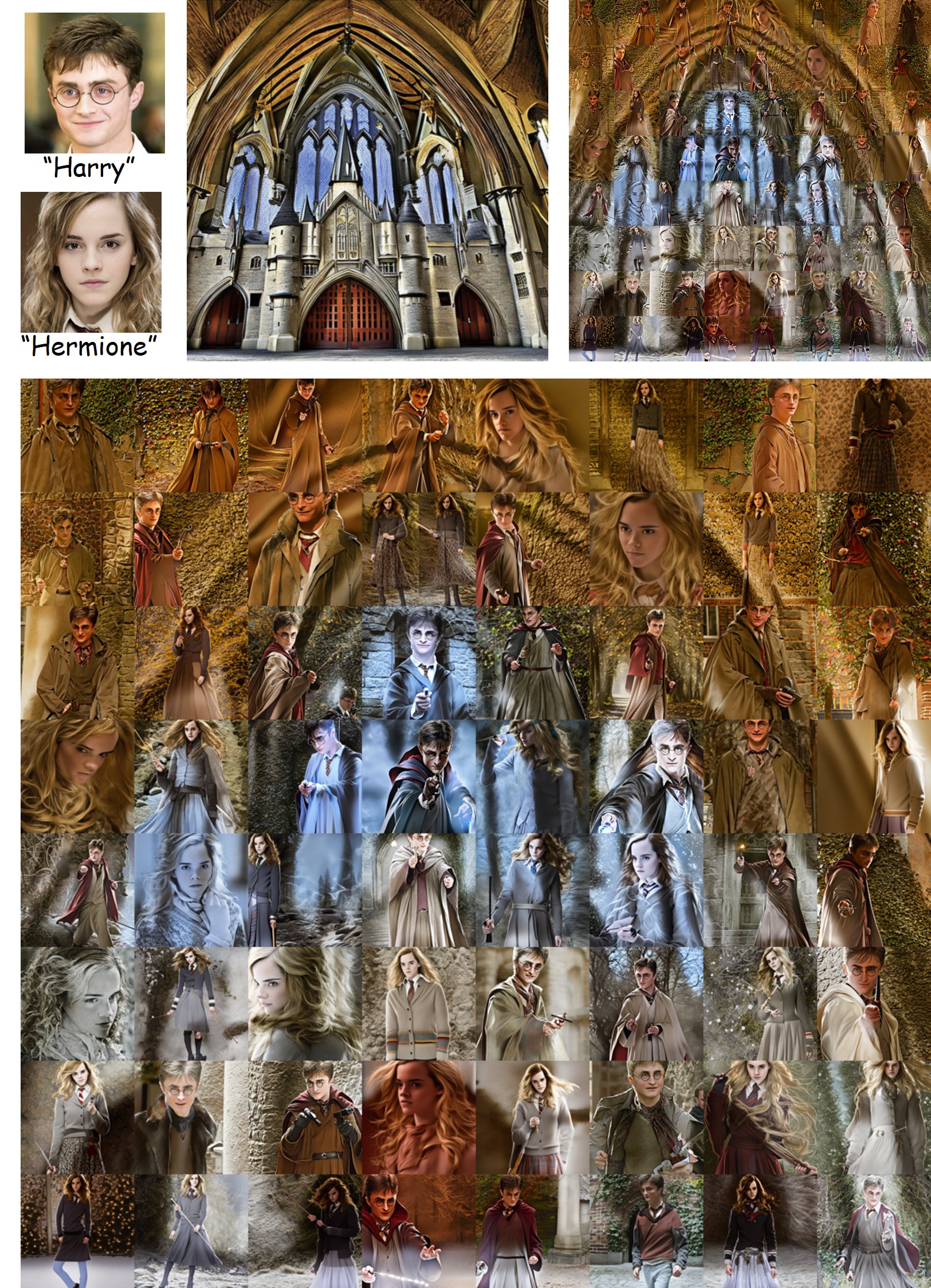}
        \caption{\centering {Personalization Example 2}: "Harry" and "Hermione"}
    \end{subfigure}
    \hfill
    \begin{subfigure}{\ww}
        \centering
        \includegraphics[width=\linewidth, height=2\linewidth, keepaspectratio]{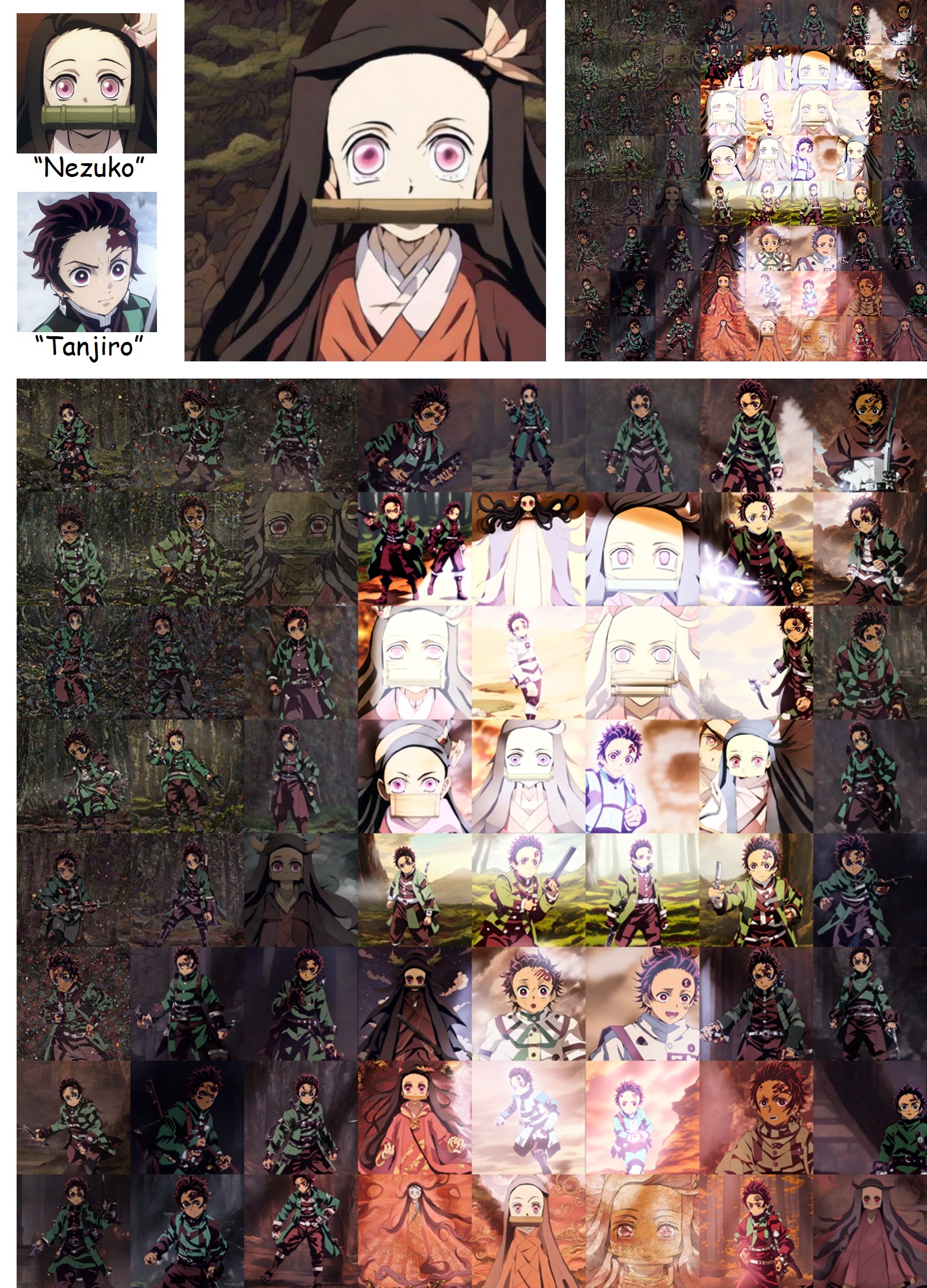}
        \caption{\centering {Personalization Example 3}: "Nezuko" and "Tanjiro"}
    \end{subfigure}
    \vspace{-1mm}

    \caption{
        \textbf{Application}: Personalized Generative Photomosaic.
    }
    \label{fig:personalization}
    \vspace{-4mm}
\end{figure*}


\subsection{Application: Personal Photomosaic Generation}
We present personalized photomosaic results in \Cref{fig:personalization}. 
To construct user-specific tile appearances, we collected 15 reference images per sample from the internet and performed LoRA fine-tuning using the Mix-of-Show \cite{gu2023mix}. 
Each ED-LoRA weight required approximately twenty minutes of training. 
During generation, the personalized LoRA weights are incorporated into the diffusion model so that each tile image reflects the target concept while still preserving the global structure of the reference image. 
As a result, the generated photomosaic simultaneously maintains global visual fidelity to the target image and local semantic consistency with the personalized concept.
In addition, the reference image itself can also be generated using the same personalized concept, enabling a fully personalized photomosaic where both the global structure and local tiles share the same identity or theme.
These results demonstrate that our framework can flexibly generate photomosaics that align with user-preferred visual targets, illustrating the adaptability of our approach to personalized content creation.

\section{Conclusion}
We presented the first generative framework for photomosaic creation, replacing traditional matching-based pipelines with diffusion-based tile synthesis that preserves global structure and supports prompt-driven detail. Through low-frequency–conditioned guidance and color distribution matching, our method produces coherent, diverse, and user-adaptive photomosaics without requiring large tile collections. The results demonstrate the effectiveness of generative modeling for photomosaic composition and open new directions for controllable image assembly, including personalized photomosaic generation as a practical downstream application.

%
%
\bibliographystyle{splncs04}
\bibliography{main}

\newpage
\appendix

\section{Implementation Details}
\subsection{Evaluation Details}
\begin{figure*}
    \vspace{-6mm}
    \centering
    \includegraphics[width=1\linewidth]{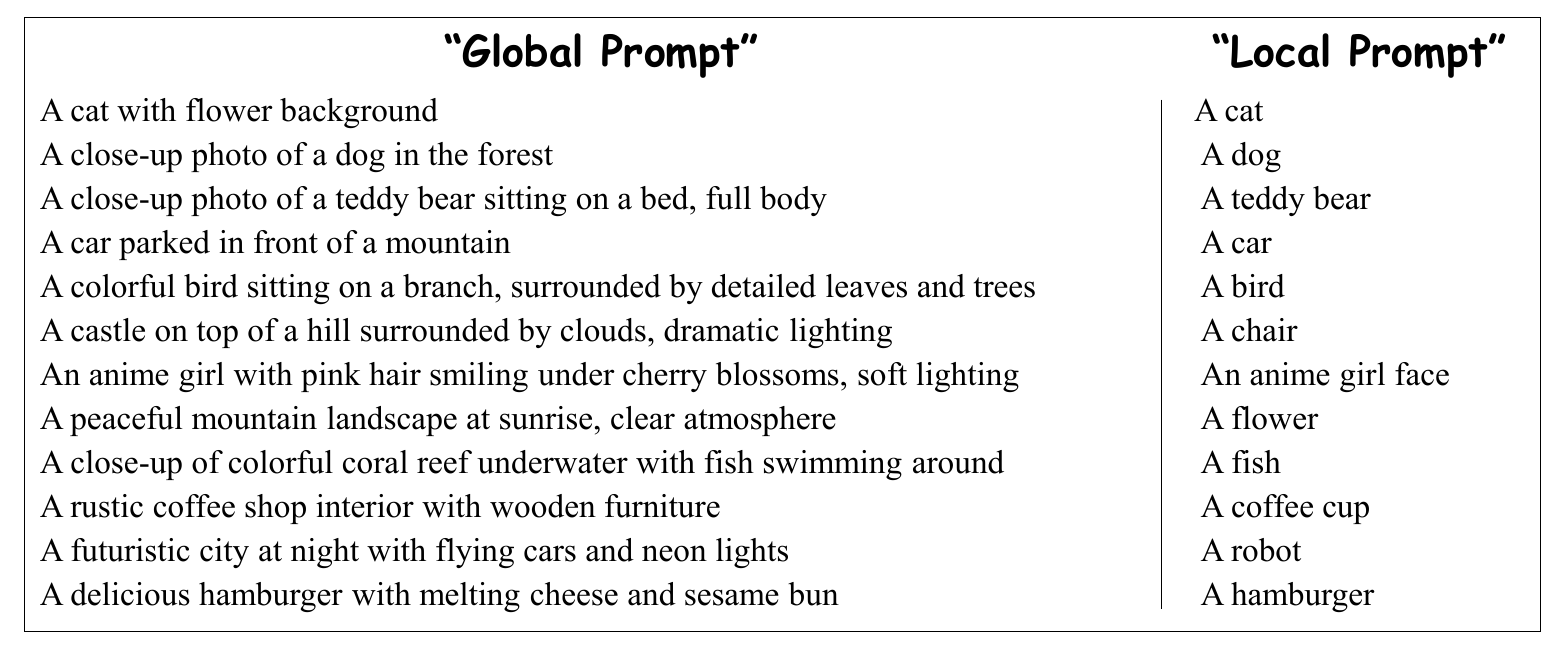}
    \vspace{-6mm}
    \captionof{figure}{\textbf{Global and Local Prompt Sets for Quantitative Evaluation.}}
    \vspace{-6mm}
    \label{fig:prompt_set}
\end{figure*}

We evaluate all methods using 12 global and local prompt sets and 15 fixed random seeds to ensure reproducibility and fair comparison of quantitative metrics.
The prompts used for photomosaic generation are shown in \Cref{fig:prompt_set}.

\section{Additional Experiments}
\subsection{Comparison to Factorized Diffusion}
\begin{figure}[h]
    \newcommand{\ww}{0.5\linewidth}
    \newcommand{\hh}{0.5\linewidth}
    \centering
    \includegraphics[width=\linewidth]{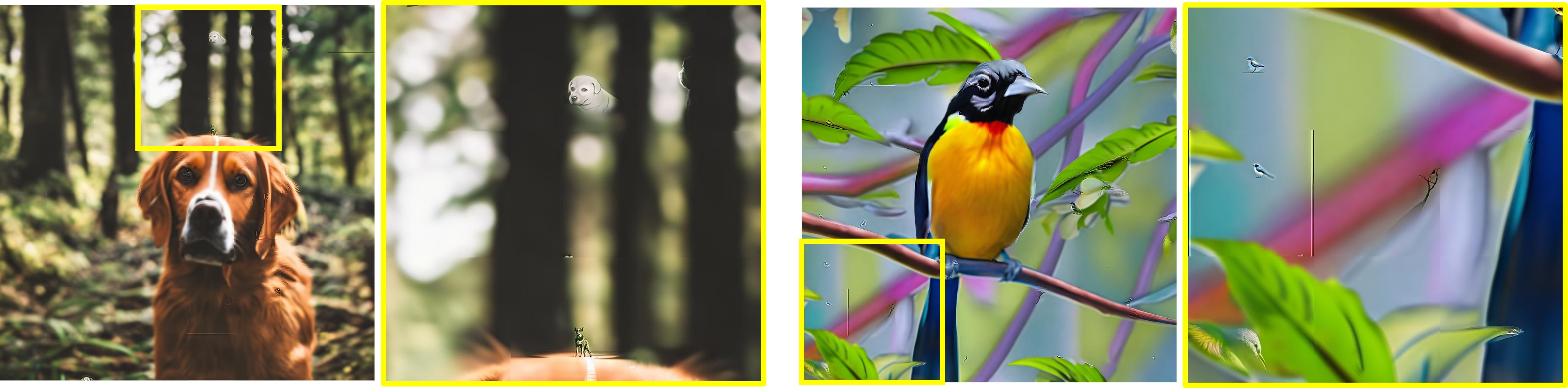} 
    \captionof{figure}{\textbf{Qualitative Result of Factorized Diffusion.}}
    \label{fig:factorized}
    \vspace{-2mm}
\end{figure}

We conduct additional experiment by applying a recent technique, factorized diffusion (FD)~\cite{geng2024factorized} 
to generate photomosaic image.
Factorized diffusion decomposes an image into multiple components and performs diffusion sampling by conditioning each component on different prompts.
This baseline relies on pixel diffusion, whereas our method and existing baselines perform diffusion in the latent space, we additionally report its results in \Cref{tab:facto} and \Cref{fig:factorized}.
We implement this baseline with its default settings within photomosaic framework.
Factorized diffusion achieves strong global structural fidelity, but exhibits weak local tile semantic alignment. 
In particular, many tiles fail to generate the intended concepts, and the resulting tile quality is often degraded.

\begin{table}[h]
    \centering

    \caption{\textbf{Comparison to Factorized Diffusion.}}
    \vspace{-2mm}
    \resizebox{0.95\linewidth}{!}{
    \begin{tabular}{c|ccccc|ccccc}
        \toprule
        & \multicolumn{5}{c|}{Ref. Image (Global)} & \multicolumn{5}{c}{Tile Image (Local)} \\
        &  PSNR$_{\uparrow}$ & SSIM$_{\uparrow}$ & LPIPS$_{\downarrow}$  & HPSv2$_{\uparrow}$ & IR$_{\uparrow}$ & BLIP$_{\uparrow}$ & CLIP$_{\uparrow}$ & IQA$_{\uparrow}$ & HPSv2$_{\uparrow}$ & IR$_{\uparrow}$ \\
        \midrule
        FD &27.007&0.937&0.010&0.134&-2.174&0.444&0.602&0.383&0.174&-1.574\\
        Ours & 17.598 & 0.551 & 0.121 &0.132 & -2.225 & 0.569 & 0.619 & 0.812 & 0.239 & -0.317  \\
        \bottomrule
    \end{tabular}
    }
    \label{tab:facto}
    \vspace{-1mm}
\end{table}

\subsection{Low-Frequency Alignment Analysis}
We evaluate global structural fidelity using a multi-scale Gaussian pyramid and measure the coarse-scale discrepancy between the reference and generated mosaic images.
We construct a pyramid by recursively applying Gaussian smoothing and $2 \times$ downsampling, producing progressively coarser representations that preserve only large-scale shape and tonal structure.
We denote these pyramid levels as $Level$-1 through $Level$-4, each representing a lower-frequency band and capturing increasingly broader structural components of the image.
Since high-frequency details are removed at coarser levels, this measure isolates global structural alignment.
For each pyramid level~$k$, we quantify the structural difference between the reference and the mosaic image using mean squared error:
\begin{equation}
\label{eq:coarse_mse}
\mathcal{E}_k= \left\|\, G_k(x) - G_k(\hat{x}) \,\right\|_2^{2},
\end{equation}
where $x$ and $\hat{x}$ denote the reference and mosaic images, and $G_k(\cdot)$ denotes the $k$-th Gaussian-pyramid level.
The metric $\mathcal{E}_k$ reflects the discrepancy in large-scale structure and tone after high-frequency components have been removed.
Lower values indicate better global structural consistency.
As shown in \Cref{tab:s_lap}, our method consistently achieves the lowest error, demonstrating superior alignment at coarse spatial scales.

\begin{table}[t]
    \centering
    \caption{\textbf{Low-Frequency Alignment Evaluation.}}
    \vspace{-2mm}
    \resizebox{0.65\linewidth}{!}{
    \begin{tabular}{c|cccc}
        \toprule
        & $Level$-1 & $Level$-2 & $Level$-3 & $Level$-4 \\
        \midrule
        Match \& Tone [15] &0.026&0.016&0.012&0.008 \\
        Color ControlNet [20] &0.107&0.088&0.072&0.054 \\
        AdaIN [10] &0.044&0.025&0.018&0.012  \\
        NoiseBlend [16] &0.067&0.014&0.008&0.005\\
        StreamDiffI2I [12] &0.025&0.016&0.013&0.011  \\
        \textbf{Ours} & \textbf{0.025} & \textbf{0.009} & \textbf{0.005} & \textbf{0.003}  \\
        \bottomrule
    \end{tabular}
    }
    \label{tab:s_lap}
\end{table}

\subsection{Computation Cost and Runtime}
We report the time required to construct a photomosaic composed of 64 tiles in \Cref{tab:recon}, measured on an NVIDIA A6000 GPU; Model loading/image pool generation time are excluded.
We employ Stable Diffusion 2.1 in our implementation. 
However, we expect that incorporating fast generation strategies based on few-step diffusion models, such as SDXL-Turbo and LCM, could enable more efficient photomosaic generation in future work.

\begin{table}[h]
    \vspace{1mm}
    \centering
    \caption{\textbf{Runtime Comparison.}}
    \vspace{-2mm}
    \scriptsize
    \setlength{\tabcolsep}{2pt}
    \begin{tabularx}{\linewidth}{l|*{7}{>{\centering\arraybackslash}X}}
        \toprule
        & M\&T & ControlNet & AdaIN & NoiseBlend & StreamDiffI2I & FD & Ours \\
        \midrule
        runtime(s) & 1.8 & 127 & 189 & 187 & 3.9 & 436 & 1088 \\
        \bottomrule
    \end{tabularx}
    \label{tab:recon}

\end{table}

\section{Limitations and Future Work}
While this study primarily addresses reference image-conditioned tile image generation, an interesting extension would be to incorporate a small set of real tile images as structural conditions for specific regions of the reference image. 
Such integration would allow the generated photomosaic to embed real imagery,  enhancing both its semantic coherence and artistic meaning.
Exploring this setup could further enable fine-grained regional control, opening opportunities for more intentional photomosaic design.

Additionally, video mosaics are often constructed using matching and tone-mapping approaches. In such cases, aligning motion becomes difficult, making the problem considerably more challenging. If a generative model could be used to create a video mosaic that accounts for both coarse structural consistency and optical flow, this could also be an interesting direction to explore.

\vspace{0.5em}
\noindent\textbf{Societal Impact}
Our generative photomosaic framework may raise several societal concerns.
First, malicious users could embed misleading or manipulative content within photomosaic structures, exploiting their ability to hide detailed local imagery inside a global scene.
Second, using personal or copyrighted images as tiles poses risks of privacy violation and unauthorized identity or content misuse.


\end{document}